\newcounter{licntr}			

\newcommand{\sprs}[1]{}

\newcommand{\stepli}{\refstepcounter{licntr}}	
\newcommand{\lival}{\thelicntr}			
\newenvironment{liout}{
		\stepli			
		\begin{list}{		
			(\lival)\hfill}{} 	
			\item 			
		}{ \end{list}}

\newenvironment{li*}{
	\stepli
	\samepage				
	\begin{trivlist} \item[] (\lival) \end{trivlist}
	\begin{trivlist}
	{\samepage \item[]}
	}{\end{trivlist}}					

\newcounter{alphcntr}		

\newlength{\astspace}				

\newcommand{\attrib}[1]{\mbox{{#1}}}	

\newcommand{\atomval}[1]{\mbox{{#1}}}	

\documentstyle[named,twocolumn,a4]{article}

\newcounter{lingex}

\def\lxam{
\vspace{-3.9mm}
\refstepcounter{lingex}
\begin{trivlist}
\item[]\mbox{}
\begin{tabular}{@{\hspace{0in}}r@{\hspace{0in}}ll@{\hspace{0in}}l}
(\thelingex)
     }
\vspace{-4.4mm}
\def\endlxam{\end{tabular}\end{trivlist}}

\def\gloss{
      \begin{tabular}[t]
            {@{\hspace{-\arraycolsep}}*{9}{l@{\ }}l@{}
            }
          }
\def\endgloss{\end{tabular}}

\def\fnlxam{
\begin{trivlist}
\item[]\mbox{}
\begin{tabular}{@{\hspace{0in}}r@{\hspace{0in}}lr@{\hspace{0in}}l}
     }
\def\endfnlxam{\end{tabular}\end{trivlist}}

\newcounter{listcounter}
\newcounter{myfootnote}
\newenvironment{items}
{\begin{list}{\labelitemi}
{\setlength{\parsep}{0.ex}\setlength{\itemsep}{0.ex}}}%
{\end{list}}

{\end{list}}

\begin{document}
\bibliographystyle{named}

\thispagestyle{empty}

\title{LEXICAL FUNCTIONS AND MACHINE TRANSLATION}

\author{Dirk Heylen, Kerry G. Maxwell and Marc Verhagen}

\date{OTS, Trans 10, 3512 JK Utrecht, Netherlands\\
      CLMT Group, Essex University, Colchester, Essex CO4 3SQ, England\\
      email: heylen@let.ruu.nl, kerry@essex.ac.uk, verhm@essex.ac.uk}

\maketitle

\begin{center}
cmp-lg/9410009
\end{center}

\noindent
This paper discusses the lexicographical concept of {\em lexical
functions} (\cite{Melcuk:Russian}) and their potential exploitation in
the development of a machine translation lexicon designed to handle
collocations. We show how lexical functions can be thought to reflect
cross-linguistic meaning concepts for collocational structures and
their translational equivalents, and therefore suggest themselves as
some kind of language-independent semantic primitives from which
translation strategies can be developed.\footnote{The research
reported in this paper was undertaken as the project ``Collocations
and the Lexicalisation of Semantic Operations'' (ET-10/75). Financial
contributions were by the Commission of the European Community,
Association Suissetra (Geneva) and Oxford University Press.}

\vspace*{-0.3cm}
\section{Description of the Problem}

Collocations present specific problems in translation, both in
human and automatic contexts.  If we take the construction {\em
heavy smoker} in English and attempt to translate it into French
and German, we find that a literal translation of {\em heavy}
yields the wrong result, since the concept expressed by the
adjective (something like \verb+`excessive'+) is translated by
{\em grand} (large) in French and {\em stark} (strong) in German. 
 We observe then that in some
sense the adjectives {\em stark}, {\em grand} and {\em heavy} are
{\em equivalent} in the collocational context, but that this is
of course not typically the case in other contexts, cf {\em
grande boite, starke Schachtel} and {\em heavy box}, where the
adjectives could hardly be viewed as equivalent.
It seems then that adjectives which are not literal translations
of one another may share meaning properties specifically in the
collocational context.

How then can we specify this special equivalence in the machine
translation dictionary?  The answer seems to lie in addressing the
concept which underlies the union of adjective and noun in these three
cases, i.e., intensification, and hence establish a single meaning
representation for the adjectives which can be viewed as an
interlingual pivot for translation.

Collocations have been studied by computational linguists in different
contexts. For instance, there is a substantial body of papers on the
{\em extraction\/} of ``frequently co-occurring words'' from corpora
using statistical methods (e.g., (\cite{choueka}),
(\cite{churchhanks:89}), (\cite{Smadja:92}) to list only a few). These authors
focus on techniques for providing material that can be used in other
processing tasks such as word sense disambiguation, information
retrieval, natural language generation and so on. Also, the {\em
use\/} of collocations in different applications has been discussed
by various authors ((\cite{McRoy:92}),
(\cite{Pustejovsky:cl}), (\cite{SmadjaMcKeown:90}) etc.). However,
collocations are not only considered useful, but also a {\em
problem\/} both in certain applications (e.g. generation,
\mbox{(\cite{Nirenburg:Lexical})}, machine translation,
(\cite{Heid:Collocations}))
and from a more theoretical point of view (e.g. (\cite{Abeille:89}),
\mbox{(\cite{KrennErbach93}))}.

We have been concerned with investigating the {\em lexical
functions} (LFs) of Mel'\v{c}uk (\cite{Melcuk:Russian}) as a candidate
interlingual device for the translation of adjectival and verbal
collocates. Our work is related to research by
(\cite{Heid:Collocations}).  In some respects it is an extension of some
of their suggestions. Our work differs from theirs in scope and also
in the exploration of various other directions.

\thispagestyle{empty}
\vspace*{-0.2cm}
\section{Representation}
\vspace*{-0.2cm}

The use we make of lexical functions as interlingual representations,
does not respect their original Mel'\v{c}ukian
interpretation. Furthermore, we have transferred them from their
context in the Meaning-Text Theory to a different theoretical setting.
We have embedded the concept in an HPSG-like grammar
theory.\footnote{Head Driven Phrase Structure grammar, see
(\cite{hpsg}), (\cite{hpsg2}). For another treatment of collocations in
HPSG, see (\cite{KrennErbach93}).} In this section we review this
operation. First we consider the features of Mel'\v{c}uk's treatment
that we have wanted to preserve.  Next we show how they have been
imported into the HPSG framework.

\vspace*{-0.2cm}
\subsection{Collocations and LFs}

In Mel'\v{c}uk's {\em Explanatory Combinatory Dictionary} (ECD, see
(\cite{Melcuk:e.a.})), expressions such as {\em une \underline{ferme}
intention, une r\'{e}sistance \underline{acharn\'{e}e}, un argument
\underline{de poids}, un bruit \underline{infernal}} and {\em
\underline{donner} une le\c{c}on, \underline{faire} un pas,
\underline{commetre} un crime} are described in the lexical
combinatorics zone.  These ``expressions plus ou moins fig\'{e}es''
will be called `collocations'. They are considered to consist of two
parts --- the {\em base\/} and the {\em collocate\/}. In the examples
above, the nouns are the bases and the adjectives and the verbs are the
collocates. The idea that all adjective collocates and all the verb
collocates share an important meaning component --- roughly
paraphrasable as {\em intense\/} and {\em do\/} respectively --- and
the fact that the adjectives and verbs are not interchangeable but are
restricted with this meaning to the accompanying nouns, is coded in
the dictionary using lexical functions (in this case {\bf Magn} and
{\bf Oper}).

Each article in the ECD describes what is called a `lexeme': a word in
some specific reading. In the {\em lexical combinatorics zone\/}, we
find a list of the lexical functions that are relevant to this
particular lexeme. Each lexical function is followed by one or more
lexemes (the result or value of the function applied to the head
word). The idea is that each combination of the argument with one of
the values of the function forms a collocation in our terminology. The
argument corresponds to the base and each value is a collocate.
The following features of this representation are important to us.

\begin{items}
\item Lexical functions are used to represent an important
syntactico-semantic relation between the base and the collocate.
\item The restricted combinatorial potential of the collocate lexeme
is accounted for by listing it at each base with which it can occur.
\end{items}

The second of these characteristics points out that the collocational
restriction is seen as a purely lexical, idiosyncratic one: all
collocations are explicitly listed.

One other aspect of collocations which we have to deal with is the
relation between the collocate lexeme and its freely occurring
counterpart. Collocate lexemes often differ in some respects from
their literal variants while sharing other properties.  Mel'\v{c}uk
deals with this by including in the ECD an entry for the free variant
and putting the collocate-specific information in the entry for the
base (with the result of the lexical functions). The full entry of the
collocate is the result of taking the entry for the free variant and
overwriting it with the information provided at the base.

\vspace*{-0.3cm}
\subsection{Collocations in HPSG}

The three aspects of Mel'\v{c}uk's analysis we wanted to encode in
HPSG were the following.

\begin{items}
\item Coding the base-collocate relation in the lexicon.
\item Choosing the level at which lexical functions will be situated.
\item Relating the collocate information to the free variant entry.
\end{items}

We have provided straightforward solutions to these problems. For the
first problem we have taken over the ECD architecture rather directly,
by creating a dedicated `collocates' field in the entry for bases
which contains all the relevant collocates.  As far as the second
problem is concerned, the obvious place to put lexical functions is in
the semantic representation provided by HPSG.  There are various
reasons for this. One is that LFs are used in the deep syntax level in
Mel'\v{c}uk's model, a level oriented towards meaning. Another reason
is that this level seems most appropriate to be used in
transfer/translation and because we want to use lexical functions in
transfer, this is where they should be. In contrast to the ECD, the
meaning of the collocate is represented by the lexical function only.

The following is an example of the entry for {\em criticism} with the
encoding of {\em strong} as a collocate.\footnote{Notice that here we
use a simple version of HPSG based on \mbox{(\cite{hpsg})} whereas the actual
implementation was based on (\cite{hpsg2}).} We use {\sc sem\_ind} as an
abbreviation for the feature path {\sc sem.cont.ind}.

\medskip

\footnotesize
\hspace*{-3mm}
$
\left[\begin{array}{ll}
\attrib{PHON} & \atomval{criticism}\\
\attrib{SEM\_IND} &
  \left[\begin{array}{ll}
  \attrib{VAR}  & \hspace*{-2mm}\atomval{\fbox{1}}\\
  \attrib{REST} & \hspace*{-2mm}\atomval{\{criticism(\fbox{1})\}}\\
  \end{array}\right]\\
\attrib{COLLS} &
   \hspace*{-3.5mm}
   \{
   \left[\begin{array}{ll}
   \attrib{\$strong} & \atomval{}\\
   \attrib{SEM\_IND}
       \left[\begin{array}{ll}
       \attrib{VAR}   & \hspace*{-2mm}\atomval{\fbox{1}}\\
       \attrib{REST}  & \hspace*{-2mm}\atomval{\{{\bf Magn}(\fbox{1})\}}\\
       \end{array}\right]
   \end{array}\hspace*{-4.5mm}\right]
   \}
\end{array}\right]
$
\normalsize
\medskip

Just as in the ECD the base contains a specific zone in which the
collocates are listed. In our case, the feature `{\sc colls}' has
a set of lexical entries as its value.

Each collocate subentry bears the value of the lexical function in its
semantics field. In this representation the lexical function is chosen
as the real semantic value of the collocate. One should read the
feature structure as specifying that the semantics of {\em strong} (as a
collocate) is the predicate {\bf Magn}(\fbox{1}).


The collocate subentry only provides partial information. In fact, it
provides only the information that is specific to the occurrence of
{\em strong} in its combination with {\em criticism}. In this case
only the semantics is given.  We further assume that the lexicon also
contains a `super-entry' which provides all the information that is
shared by all the different occurrences of {\em strong}. This entry is
where the variable \$strong points to. Of course, other
architectures that try to avoid redundant specification of information
are equally possible. For instance if one assumes a mechanism of
default unification, one can have \$strong refer to the full entry
describing `strong' in say its ordinary use, and have the values that
are particular to the collocational {\em strong} overwrite the values
provided in the ordinary entry, as in Mel'\v{c}uk's proposal.

\paragraph{Collocations, Rules and Principles}

So far, we have not specified in what way one gets from the lexical
entries for the base and the collocate to the representation of the
collocational expression.

In HPSG, the descriptions of complex expressions are constrained by
principles. We will assume that collocations are subject to the same
constraints. The ordinary rules of combination (combining adjectives
and nouns, for instance) thus account for most of the properties of
the collocational combination. However, we are still left with the
typical `collocational restriction' which needs to be accounted for.

We have therefore added a principle which says that constructions that
are analysed as collocations (indicated by the type {\sc collocation})
are either head-adjunct structure or head-complement structures with
specific restrictions holding between the head and the adjunct or the
head and the complement respectively. Let's consider the former
case\footnote{To illustrate the case of head-complement structures one could
take some support verb construction (also called light verb
construction).}, illustrated by the {\em heavy smoker\/} example. The
adjunct daughter will contain the adjective collocate. In such
collocational constructions the collocate adjuncts have to be
`licensed' by the noun or the head daughter. This is implemented by
requiring that the collocates field ({\sc colls}) of the head daughter
contains a reference to a lexical entry that is compatible with the
adjunct daughter. In the literal reading of an expression such as {\em
heavy smoker\/}, the phrase will not be analysed as a {\sc
collocation} and the principle does not apply.

\medskip
\footnotesize
\hspace*{-1mm}

$\left[\begin{array}{ll}
\\
\end{array}\right]_{\mbox{COLLOCATION}} \Rightarrow$

$
\begin{array}{c}

\left[\begin{array}{ll}
\attrib{HEAD\_DTR} & \left[\begin{array}{ll}
           \atomval{COLLS} & \{...\fbox{1}...\}
           \end{array}\right]\\
\attrib{ADJ\_DTRS} & < ...\fbox{1}_{\mbox{COLLOCATE}}...>
\end{array}\right]\\

\bigvee\\

\left[\begin{array}{ll}
\attrib{HEAD\_DTR} & \fbox{1}_{\mbox{COLLOCATE}}\\
\attrib{COMP\_DTRS}& <...[\attrib{COLLS} \{...\fbox{1}...\}]...>
\end{array}\right]
\end{array}$

\normalsize
\medskip

\vspace*{-0.3cm}
\section{Issues in Translation}
\label{translation}

\sprs{Despite their differences, the classical MT {\em transfer\/} and {\em
interlingua\/} architectures share the same basic mechanics. Both fall
within the paradigm that is concerned with mappings between symbolic
representations that start and end with natural language. Differences
between these architectures and their variants can be characterised by
the number of representation levels, their interpretation, and the
mappings between them.

Within linguistics and MT the following levels of representation and
other aspects of linguistic utterances are often distinguished:
phonology, syntax, semantics, pragmatics, extra-linguistic knowledge.}

The project has tried to investigate the use of lexical functions as
an interlingual device, i.e., one which is shared by the semantic
representations of collocations in the language pairs\footnote{For
another application of LFs in a multilingual NLP context see
(\cite{Heid:Collocations}). For other treatments of collocations in language
generation see (\cite{Nirenburg:Lexical}) and (\cite{SmadjaMcKeown:90}).}.

The typing of a collocation with such a function opens up the way to a
treatment of collocations inside a given language module and hence to
a substantial reduction in the number of collocations explicitly
handled in the multilingual transfer dictionary. The existence
of a collocation function is established during analysis. This
information is used to generate the correct translation in the target
language. To illustrate, the English analysis module might analyse (1)
as (2). The transfer module maps (2) onto (3) which is then
synthesised by the French module to (4).

\begin{quote}
(1) heavy smoker $\rightarrow$ (2) {\bf Magn}(smoker)
$\rightarrow$ (3) {\bf Magn}(fumeur) $\rightarrow$ (4) grand fumeur
\end{quote}

The example points out that the translation strategy is a mixture of
transfer and interlingua. The bases are transferred but the
representation of the collocate is shared between the source and the
target representation. This treatment of collocations rests, among
others, on the assumptions that there are only a limited number of
lexical functions, that lexical functions can be assigned consistently,
that all (or a significant number of) collocations realise a lexical
function, that lexical functions are not restricted to particular
languages, etc. In the following paragraph we present an outline of the
translation process. Next, we discuss some of the problems which
follow from our approach and we propose some ways to solve them.

\vspace*{0.2cm}
\subsection{Lexical Functions as Interlingua}

It was assumed that the starting point for transfer is the semantic
representation of the phrase. Using a semantic representation as input
to transfer implies that we relate semantic values of words and
phrases. For our purposes this is very satisfying since we will now be
using the semantics of collocates instead of their orthography, in
other words: we use lexical functions and abstract away from the
particular realisation of a collocate in a particular language.

We now state the relation between the semantic representations of the
source language and target language. The semantic relation between the
phrase {\em heavy smoker} and its French counterpart can be made
explicit in the following bilingual sign:

\sprs{
\medskip
\begin{figure}[h]
\tiny
$
\left[\begin{array}{ll}
\attrib{EN|SEM|CONT|IND} &
   \left[\begin{array}{ll}
      \attrib{VAR} & \atomval{\fbox{1}}\\
      \attrib{REST} &
         \{
         \left[\begin{array}{ll}
         \attrib{RELN} & \atomval{SMOKER}\\
         \attrib{INST} & \atomval{\fbox{1}}
         \end{array}\right],
         \left[\begin{array}{ll}
         \attrib{RELN} & \atomval{\bf Magn}\\
         \attrib{INST} & \atomval{\fbox{1}}
         \end{array}\right]
         \}
   \end{array}\right]\\\\
\attrib{FR|SEM|CONT|IND} &
   \left[\begin{array}{ll}
      \attrib{VAR} & \atomval{\fbox{1}}\\
      \attrib{REST} &
         \{
         \left[\begin{array}{ll}
         \attrib{RELN} & \atomval{FUMEUR}\\
         \attrib{INST} & \atomval{\fbox{1}}
         \end{array}\right],
         \left[\begin{array}{ll}
         \attrib{RELN} & \atomval{\bf Magn}\\
         \attrib{INST} & \atomval{\fbox{1}}
         \end{array}\right]
         \}
   \end{array}\right]\\
\end{array}\right]
$
\normalsize
\caption{hoi}
\end{figure}
}

\medskip
\footnotesize
\hspace*{-3mm}
$
\left[\begin{array}{ll}
\attrib{EN|SEM\_IND} &
   \hspace*{-3mm}
   \left[\begin{array}{ll}
      \attrib{VAR}  & \hspace*{-1mm}\atomval{\fbox{1}}\\
      \attrib{REST} & \hspace*{-1mm}\atomval{\{smoker(\fbox{1}),{\bf
Magn}(\fbox{1})\}}\\
   \end{array}\right]\\\\
\attrib{FR|SEM\_IND} &
   \hspace*{-3mm}
   \left[\begin{array}{ll}
      \attrib{VAR}  & \hspace*{-1mm}\atomval{\fbox{1}}\\
      \attrib{REST} & \hspace*{-1mm}\atomval{\{fumeur(\fbox{1}),{\bf
Magn}(\fbox{1})\}}\\
   \end{array}\right]\\
\end{array}\right]
$
\normalsize
\medskip


Typically, the lexicon will contain a bilingual sign for each possible
value of {\sc reln}.  Thus, for translating {\em heavy smoker} into
{\em grand fumeur} we will need the obvious entry for {\em
smoker-fumeur} plus the entry below:

\medskip
\footnotesize
\hspace*{3mm}
$
\left[\begin{array}{ll}
\attrib{EN|SEM\_IND} &
   \hspace*{-3mm}
   \left[\begin{array}{ll}
      \attrib{VAR}  & \hspace*{-1mm}\atomval{\fbox{1}}\\
      \attrib{REST} & \hspace*{-1mm}\atomval{\{{\bf Magn}(\fbox{1})\}}\\
   \end{array}\right]\\\\
\attrib{FR|SEM\_IND} &
   \hspace*{-3mm}
   \left[\begin{array}{ll}
      \attrib{VAR}  & \hspace*{-1mm}\atomval{\fbox{1}}\\
      \attrib{REST} & \hspace*{-1mm}\atomval{\{{\bf Magn}(\fbox{1})\}}\\
   \end{array}\right]\\
\end{array}\right]
$
\normalsize
\medskip

The interlingual status of the lexical function is self-evident. Any
occurrence of {\bf Magn} will be left intact during transfer and it
will be the generation component that ultimately assigns a monolingual
lexical entry to the LF.\footnote{For more details we refer the
reader to (\cite{heylen}). There we also discuss our implementation in
Alep, the C.E.C.'s unification-based grammar writing environment.}

\sprs{
Note that the bilingual entries as such do
not give sufficient information on how the bilingual sign of the whole
phrase will be constructed. Our aim here was to show the interlingual
status of lexical functions in transfer. Therefore, we abstracted away
from implementational details\footnote{For more details we refer the
reader to (\cite{heylen}). There we also discuss our implementation in
Alep, the C.E.C.'s unification-based grammar writing environment.}.}

\subsection{Problems}

Lexical Functions abstract away from certain nuances in meaning and from
different syntactic realizations. We discuss some of the problems
raised by this abstraction in this section.

\vspace*{-0.3cm}
\paragraph{Overgenerality}
An important problem stems from the interpretation of LFs implied by
their use as an interlingua --- namely that {\em the meaning of the
collocate in some ways reduces to the meaning implied by the lexical
function}. This interpretation is trouble-free if we assume that LFs
always deliver unique values; unfortunately cases to the contrary can
be readily observed.  An example attested from our corpus was the
range of adverbial constructions possible with the verbal head {\em
oppose}: {\em adamantly, bitterly, consistently, steadfastly,
strongly, vehemently, vigorously, deeply, resolutely}, etc.  The
function {\bf Magn} is an appropriate descriptor in all cases since
each adverb functions as a typical intensifier in this context.
However each adverb also denotes some other meaning aspect(s). The
imprecision of LFs will mean that we have no means of distinguishing
between the various intensifiers possible in the context of a given
keyword, and hence will not have sufficient information to choose the
most appropriate translation where, correspondingly, multiple
possibilities exist in the target language. An important question here
is how dramatic this loss of translation quality really is.

It is essentially in addressing the issue of overgenerality that
Mel'\v{c}uk introduces sub- and superscripts to lexical functions,
enhancing their precision and making them sensitive to meaning aspects
of the lexical items over which they operate.  Superscripts are
intended to make the meaning of the LF more precise and hence more
likely to imply unary mappings between arguments and values,
subscripts are used to reference a particular semantic component of a
keyword.  The introduction of such devices into the account of LFs
demonstrates both the need for precision and the fact that it does
seem necessary to address semantic aspects of lexemes standing in
co-occurrence relations.  In fact it has been asserted by some (e.g.,
(\cite{Anick}), (\cite{Heid:Collocations})) that collocational systems are
systematically predictable from the lexical semantics of nouns. In an
attempt to explore this notion further, we have investigated the
approach to nominal semantics known as {\em Qualia} structure
(\cite{Pust:91}) and considered how this may complement the LF
notion to improve its descriptive power\footnote{For a comparison
between aspects of Qualia structures and lexical functions see
(\cite{Heylen:LS}).}. Among the promising avenues that occur to us are,
firstly, the postulation of LF subscripts based on the four Qualia
roles (assuming that these are the lexically most relevant aspects of
noun semantics) and, secondly, the application of LFs to semantic
(Qualia) structures rather than monolithic lexemes; eg: the LF {\bf
Bon} is used in delivering evaluative qualifiers which are standard
expressions of praise or approval. One could imagine application of the
function over the Constitutive and Agentive roles of the noun {\em
lecture}, to deliver:

\begin{quote}
 {\bf Bon}(\verb+Const:+ lecture) = informative\\
 {\bf Bon}(\verb+Agent:+ lecture) = clear
\end{quote}

In both cases the idea is that the precision of the lexical
function is essentially enhanced by appealing to the semantic
facets of its argument.

\vspace*{-0.3cm}
\paragraph {Syntactic Divergences}

Another issue that has to be raised concerns the translation of
collocations into non-collocational constructions.  If we are to
maintain a consistent interlingual approach to the translation of
these cases, we must extend our LF-based approach accordingly. We
consider one case briefly.

Cross-linguistic analysis reveals many cases where nominal-based
collocational constructs are realised as compounds
in Germanic languages, e.g., {\em bunch of keys} $\Rightarrow$ {\em
sleutelbos}.
A possible account of such phenomena
may be developed from the concept of {\em merged\/} LFs
\mbox{(\cite{melcuk:ta})}. Merged LFs are intended to be used in cases
where a value lexeme exists which appears to effectively reduce
(``merge'') an LF meaning and its specified argument to a single
lexicalised form, rather than projecting a syntagmatic unit.  We could
argue that in cases of compound formation, exactly the same process is
to be accounted for, since the compound embodies both the concept
mediated by the LF and its argument lexeme. We could therefore allow
compounds to be delivered as values of merged LF's, eg:
{\bf //Mult}(sleutel)= sleutelbos.

These observations are useful in the MT context if we assume that we
can effect a mapping between merged and unmerged LFs and therefore
capture the correspondence between distinct structural realisations of
the same concept.  One way to emulate such a mapping might be through
the use of Mel'\v{c}uk's {\em lexical paraphrasing rules}. For
instance, one could conceive of a lexical paraphrasing rule as
follows\footnote{This is our own initiative -- it seems to be the
case as we examine the literature that neither LFs such as {\bf
Magn}, {\bf Bon} etc (i.e., those representing standard
qualifiers/attributes) nor indeed {\em merged} LFs feature in lexical
paraphrasing rules. We would argue that cross-linguistic analysis
suggests that they should enter this domain; compound formation and
other types of lexicalisation appear to be regular patterns of
translation across many collocational constructs, as we illustrate
here.}:
$  W + \mbox{{\bf Mult}}(W)
   \Longleftrightarrow
   \mbox{{\bf //Mult}}(W) $.

If we assume that in our
monolingual English lexicon, we assign the collocate {\em bunch}
as the {\bf Mult} value of keyword {\em key}, and that
accordingly in the Dutch lexical entry for {\em sleutel} we
instantiate {\em sleutelbos} as the value of the merged LF {\bf
//Mult}, then we can use the paraphrasing rule to effect a mapping
between the two LF's and hence arrive at an interlingual approach
to the translation of the example, despite structural mismatches, i.e.,
\begin{center}
key + bunch[$\mbox{\bf Mult}(key)$]

$\Longleftrightarrow$

sleutelbos[$\mbox{\bf //Mult}(sleutel)$]
\end{center}

Further examples exist where productive morphological processes (e.g.,
affixation\footnote{One could think of an example such as {\em
mis-interpret\/}.}) lead to the lexicalisation in one language of
concepts that exist as syntagmatic constructs in another.  Again, we
suggest the use of merged LFs and corresponding mappings via lexical
paraphrasing rules as a possible translation strategy in these cases.

\section{Summary and Conclusions}

In this paper we have discussed how the lexicographical concept of
{\em lexical functions}, introduced by Mel'\v{c}uk to describe
collocations, can be used as an interlingual device in the machine
translation of such structures. We have shown how the essentials of
the ECD analysis can be embedded in the lexicon and grammar of a
unification based theory of language.

Our use of lexical functions as an interlingua assumes that the
relevant aspects of the meaning of the collocate are fully
captured by the LF. The LF therefore determines the accuracy of
translations, which may be impoverished due to the generalised
nature of basic LFs. We have suggested some ways in which LFs
can be enriched with lexical semantic information to improve
translation quality.

The interlingua level reflects what is semantically common to
expressions which form translational equivalents. It abstracts
away from specific syntactic realisations. Given that
collocations may translate as non-collocations, we also have to
provide a way to represent these expressions using lexical
functions. We have provided an illustration on how to
proceed in one such case.

\paragraph{Acknowledgements} We would like to thank the following
partners and colleagues: Susan Armstrong-Warwick, Laura Bloksma,
Nicoletta Calzolari, R. Lee Humphreys, Simon Murison-Bowie and
Andr\'{e} Schenk.





\end{document}